\documentclass[sigconf]{acmart}

\copyrightyear{2024}
\acmYear{2024}
\setcopyright{acmlicensed}\acmConference[SIGIR '24]{Proceedings of the 47th International ACM SIGIR Conference on Research and Development in Information Retrieval}{July 14--18, 2024}{Washington, DC, USA}
\acmBooktitle{Proceedings of the 47th International ACM SIGIR Conference on Research and Development in Information Retrieval (SIGIR '24), July 14--18, 2024, Washington, DC, USA}
\acmDOI{10.1145/3626772.3657978}
\acmISBN{979-8-4007-0431-4/24/07}
\acmPrice{}
\usepackage{graphicx}            %
\usepackage{duckuments}          %

\usepackage[utf8]{inputenc}
\usepackage{longtable}
\usepackage{threeparttable}
\usepackage{xspace}
\usepackage{appendix}
\usepackage{booktabs}
\usepackage[utf8]{inputenc}
\usepackage[
 font=small %
 ]{subfig}
\usepackage{textcomp}
\usepackage{amsmath, amsthm}
\usepackage{amsfonts}
\usepackage{enumitem}
\usepackage{algorithm}
\usepackage[noend]{algpseudocode}
\algtext*{EndProcedure}%
\usepackage{xcolor,colortbl}
\usepackage{footnote}
\usepackage{url}
\usepackage{tabularx}
\usepackage{multicol}
\usepackage{multirow}
\usepackage{pifont}
\usepackage{bm}
\usepackage{listings}
\usepackage{wrapfig}
\usepackage{relsize}
\usepackage{mathtools}
\usepackage{bbm}
\usepackage{dsfont}
\usepackage{blindtext}
\usepackage{bm}
\usepackage{balance}
\usepackage{mdwlist}
\usepackage{paralist}
\usepackage{diagbox}
\usepackage{tikz}
\setlist[enumerate]{leftmargin=*}
\setlist[itemize]{leftmargin=*}
\usepackage[customcolors]{hf-tikz}
\usepackage[hang,flushmargin]{footmisc}

\usepackage{wasysym}
\usepackage{pifont}%

\definecolor{green}{rgb}{0,0.7,0.3}
\definecolor{msblue}{rgb}{0.310,0.506,0.741}
\definecolor{msorange}{rgb}{0.968,0.588,0.274}
\definecolor{msgray}{rgb}{0.5,0.5,0.5}

\newcommand{\tikzsquare}[2][red,fill=red]{\tikz[baseline=-0.5ex]\draw[#1,radius=#2] (-0.1,-0.1) rectangle (0.13,0.13) ;}%

\newcommand{\method}{\textsc{TouchUp-G}\xspace}

\newtheorem{definition}{Definition}

\newcommand{\myTag}[1]{{\textbf{#1}}}
\newcommand{\failCell}{\xmark}
\renewcommand{\failCell}{}
\newcommand{\winCell}{{\color{green}{\cmark}}}
\newcommand{\general}{{\bf General}}
\newcommand{\multimodal}{{\bf Multi-modal}}
\newcommand{\principled}{{\bf Principled}}
\newcommand{\effective}{{\bf Effective}}

\newcommand{\matA}{\mathbf{A}}

\newcommand{\matG}{{G}}

\newcommand{\matV}{V}
\newcommand{\matX}{\mathbf{X}}

\newcommand{\matT}{\mathbf{T}}

\newcommand{\matE}{{E}}
\newcommand{\feat}{\textbf{$x$}}

\newcommand{\ogbproducts}{\texttt{Ogb-Products}}
\newcommand{\books}{\texttt{Books}}
\newcommand{\ogbarxiv}{\texttt{Ogb-Arxiv}}
\newcommand{\amazoncp}{\texttt{Amazon-CP}}

\newcommand{\matS}{{S}}

\newcommand{\cmark}{\ding{51}}%

\def\etal{\emph{et al.}}

\newcommand{\bert}{BERT+}
\newcommand{\degree}{Degree+}
\newcommand{\deepwalk}{Deepwalk+}
\newcommand{\vit}{ViT+}
\newcommand{\ogb}{Ogb+}
\newcommand{\deberta}{DeBERTa+}
\newcommand{\scibert}{SciBERT+}

\newcommand{\matrices}{Feature Homophily\xspace}

\newcounter{theorem}

\definecolor{Gray}{gray}{0.85}

\tikzset{offset def/.style={
        above left offset={-0.2,0.5},
        below right offset={0.35,-0.3},
    },
    color def/.style={
        offset def,
        set fill color=white,
        set border color=red,
    },
}

\author{Jing Zhu*}
\affiliation{\institution{University of Michigan, Ann Arbor} \country{Ann Arbor, MI, USA}\thanks{* Work done during an internship at Amazon} }
\email{jingzhuu@umich.edu}

\author{Xiang Song}
\affiliation{\institution{Amazon} \country{Santa Clara, CA, USA}}
\email{xiangsx@amazon.com}

\author{Vassilis N. Ioannidis}
\affiliation{\institution{Amazon} \country{Santa Clara, CA, USA}}
\email{ivasilei@amazon.com}

\author{Danai Koutra}
\affiliation{\institution{University of Michigan, Ann Arbor}\country{Ann Arbor, MI, USA}}
\email{dkoutra@umich.edu}

\author{Christos Faloutsos}
\affiliation{\institution{Carnegie Mellon University}\country{Pittsburgh, PA, USA}}
\email{christos@cs.cmu.edu}

\begin{document}

\begin{abstract}

How can we enhance the node features acquired from Pretrained Models (PMs) to better suit downstream graph learning tasks? Graph Neural Networks (GNNs) have become the state-of-the-art approach for many high-impact, real-world graph applications. 
For feature-rich graphs, a prevalent practice involves directly utilizing a PM to generate features. Nevertheless, this practice is suboptimal as the node features extracted from PMs are graph-agnostic and prevent GNNs from fully utilizing the potential correlations between the graph structure and node features, leading to a decline in GNN performance.
In this work, we seek to improve the node features obtained from a PM for graph tasks and introduce \method, a "Detect \& Correct" approach
for refining node
features extracted
from PMs. \method \textbf{detects} the alignment using a novel feature homophily metric and \textbf{corrects} the misalignment through a simple \underline{touchup} on the PM.
It is
(a)~\general: applicable to any downstream graph task; 
(b)~\multimodal: able to improve raw features of any modality; 
(c)~\principled: it is closely related to a novel metric, feature homophily,  which we propose to quantify the alignment between the graph structure and node features; 
(d)~\effective: achieving state-of-the-art results on four real-world datasets spanning different tasks and modalities.

\end{abstract}

\ccsdesc[500]{Information systems~Information retrieval}

\keywords{graph-centric finetuning; feature homophily; graph neural network}

\title{\method: Improving Feature Representation through Graph-Centric Finetuning} %

\maketitle

\section{Introduction}
\label{sec:intro}

\vspace{-0.2em}
 Graphs or networks serve as fundamental representations for relational structures, and their analysis is useful in many scientific and industrial applications. Various tasks, ranging from recommendation and molecule property prediction to knowledge graph completion, can be formulated as graph learning endeavors.  
\begin{figure}[t!]
	\centering
    \vspace{-0.7cm}
	\setcounter{subfigure}{0}
	
    \subfloat[Quantitative Result\label{fig:teaser-quan}]{%
      \includegraphics[width=0.23\textwidth]{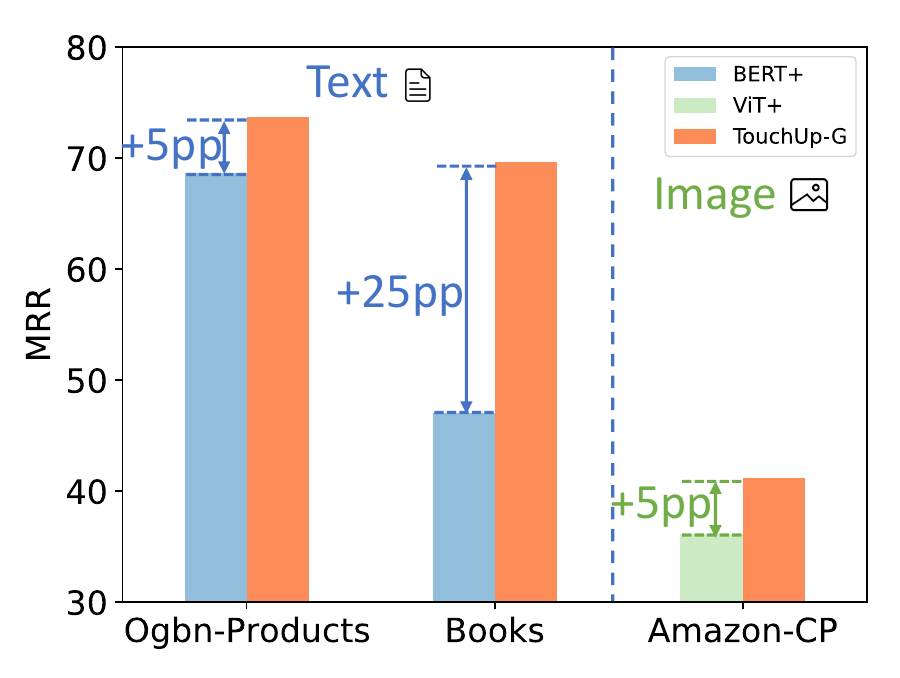}
    }  
    ~
    \subfloat[Qualitative Result\label{fig:teaser-qual}]{%
      \includegraphics[width=0.23\textwidth]{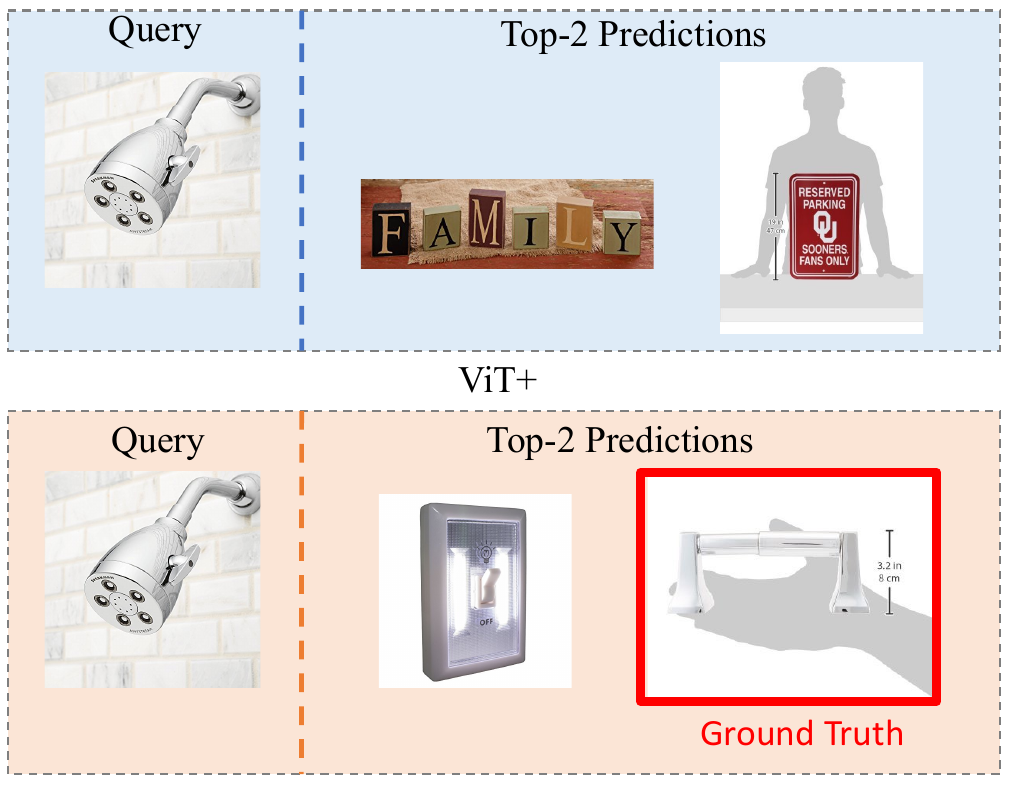}
    }

\vspace{-0.2cm}
    \caption{\myTag{\method wins:}
(a) Compared with features obtained directly from PMs (BERT~\cite{devlin2018bert} or ViT~\cite{dosovitskiy2020image}), \method  improves the quantitative performance by more than 25$\%$ across datasets and modalities. (b)
Examples from the Amazon co-purchasing graph (\amazoncp) show that \method correctly predicts the ground truth while \vit ~fails.}
\vspace{-0.5cm}
    \label{fig:teaser}
\end{figure}

Pretrained Models (PMs), such as BERT, GPT, and ViT~\cite{devlin2018bert, brown2020language, dosovitskiy2020image} have demonstrated remarkable performance across a range of natural language and computer vision tasks, including question answering and image classification. These models have become the foundations of modern ML systems. In real-world applications, it is common practice to utilize pretrained models for generating node features, and subsequently integrating these derived features into GNNs for graph learning tasks, encompassing link prediction and node classification~\cite{jin2022heterformer,hu2021ogb, hu2020open, zhao2022learning, chien2021node}. Nonetheless, employing node features from PMs without any domain adaptation is suboptimal, as features generated from PMs are  unaware of the graph structure and prevents GNNs from fully exploiting the relation between node features and the graph structure.

\begin{wrapfigure}{r}{0.30\textwidth}
\vspace{-0.4cm}
\centering
\includegraphics[width=0.22\textwidth]{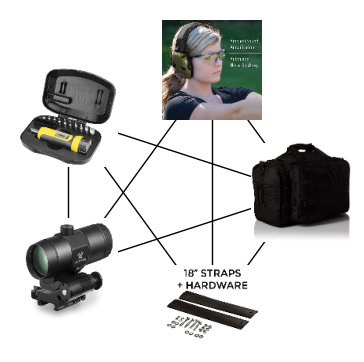}
\vspace{-0.5cm}
\caption{\myTag{Why pretrained features may fail:} We show a subgraph of Amazon Co-purchasing graph (Amazon-CP). 
Products possessing disparate visual features are often bought together.
}
    \label{fig:amazon-cp}
    \vspace{-0.5cm}
\end{wrapfigure}

An illustrative real-world example is shown in Fig.~\ref{fig:amazon-cp}, where we leverage a co-purchasing graph to predict products frequently purchased together. Utilizing image features directly from PMs results in distinct feature representations for each product, causing GNNs to falter in predicting products that tend to be purchased together. This is a widespread issue that persists whenever a PM is leveraged to generate node features and the pretraining objective does not take graph structure into account ~\cite{chien2021node}. 
Existing methods to overcome this issue suffer from two drawbacks. First, they do not measure the quality of node features from PMs, which decides whether finetuning is necessary. Second, they use text-specific features and only work on text-rich graphs.
To address these issues, we introduce a \textit{principled} and \textit{general} solution, which improves node features obtained from \emph{any} PM, so that GNNs can achieve better performance in \emph{any} downstream graph task.  

Specifically, we propose \method, a "Detect \& Correct" approach for refining node features extracted from PMs. To \textbf{detect} the alignment between the node features and the graph structure, we introduce a novel feature homophily metric that quantifies the potential correlations between graph structure and node features. Our proposed feature homophily metric accommodates vectorized features and is bounded in the range of [-1,1] so as to allow comparison between graphs.
To \textbf{correct} the node features obtained from PMs, we propose graph-centric finetuning, a simple \underline{\textsc{TouchUp}} enhancement technique that improves \underline{G}raph's node features from any PM. 

\method is simple, adaptable to a variety of PMs in multiple domains. The node features obtained from \method show a strong alignment with the graph structure and attain state-of-the-art performance for downstream graph tasks. A summary of our results is shown in Fig.~\ref{fig:teaser}.
Our contributions are summarized as follows: 

\begin{compactitem}
    \item \general:\method can be applied to various graph tasks. 
    \item \multimodal: \method can be applied to any PM from any modality, e.g., texts, images. To the best of our knowledge, we are the first to propose finetuning vision transformers for graph tasks. 
    \item \principled: We propose a novel metric called feature homophily, (Eq.~(~\ref{eq:homophily}), \S~\ref{sec:homophily}), to measure the correlation between node features and graph structure.
    \item \effective: \method achieves state-of-the-art performance on four real datasets across various tasks and modalities. 
\end{compactitem}

\section{Related Work}
\label{sec:related}
\begin{table}[t]
\caption{Qualitative comparison of finetuning frameworks. 
\method is \general--it can be
applied on any graph tasks--, \multimodal\xspace  (applicable to features from any modality), \principled \xspace and \effective.
\label{tab:salesman}}
\centering
\vspace{-0.4cm}
\resizebox{0.9\columnwidth}{!}{
\begin{tabular}{l|cccc||c}
\toprule
  \diagbox{Property}{Method} & \rotatebox[origin=c]{0}{GLEM ~\cite{zhao2022learning}}
  &  \rotatebox[origin=c]{0}{GIANT ~\cite{chien2021node}} 
  &  \rotatebox[origin=c]{0}{Patton ~\cite{jin2023patton}}
  &  \rotatebox[origin=c]{0}{BERT+ ~\cite{devlin2018bert}}&
  \rotatebox[origin=c]{0}{\textbf{\method} (ours)} \\  
\midrule
  \textbf{\general-NC} & \winCell & \winCell  & \winCell & \winCell & \winCell \\

  \textbf{\general-LP} & \failCell & \failCell  & \winCell & \winCell & \winCell \\
  \textbf{\multimodal} & \failCell & \failCell  & \failCell   & \failCell  & \winCell\\

   \textbf{\principled} & \failCell &\winCell  & \failCell  & \failCell & \winCell\\

   \textbf{\effective} & \winCell& \winCell  & \winCell & \textbf{?} &  \winCell\\
\bottomrule
\end{tabular}
}
\vspace{-0.3cm}
\end{table}

\noindent \textbf{Pretrained Models as Feature Embeddings.} 
Pretrained models, have been widely used to acquire broad language and visual representations~\cite{devlin2018bert, liu2019roberta, liu2023pre, yasunaga2022linkbert, vaswani2017attention, liu2021swin,kirillov2023segment,dosovitskiy2020image,ranftl2021vision}. For transformers in the language and vision domain, we refer to the survey for more details ~\cite{khan2022transformers}. 
The typical approach of learning text-rich graphs adopts
a “cascaded architecture”  where features are extracted via a PM and incorporated into GNNs. But this approach is suboptimal and several works have been proposed to better adapt features from PMs to GNNs ~\cite{ioannidis2022efficient, chien2021node, zhao2022learning, jin2023patton, xie2023graph}. 
A comparative analysis between \method and previous methods is outlined Table~\ref{tab:salesman}. Here, we aim to design a \textit{general} approach that works for any source of modality, such as images, and any graph downstream task. Moreover, we measure how misaligned the node features and graph structure are, which is important to decide whether enhancement is needed.

\noindent \textbf{Feature Homophily in GNNs.} Existing GNN models heavily rely on the feature homophily assumption. Yoo ~\etal~highlight the inability of GNN models to effectively leverage the graph structure in the presence of noisy node features, primarily due to their strong dependence on the feature matrix~\cite{yoo2023less}. Yang \etal \xspace show that GNNs penalize deviations between the embeddings of two nodes sharing an edge ~\cite{yang2021graph, yang2021implicit}. However, what is a good metric for feature homophily remains unclear. Newman 
introduced assortativity as a measure to assess the similarity of scalar features along  edges~\cite{newman2002assortative, newman2003mixing}, but it cannot be applied to vectorized features. Feature smoothness is another candidate metric~\cite{hou2022measuring}, but it is unbounded in magnitude, making it hard to compare against graphs with varying sizes. In contrast, our proposed feature homophily score accommodates vectorized features and allows comparison across graphs of different sizes.

\section{\method}
\label{sec:method}

\subsection{Preliminaries}

\vspace{-0.1cm}
\noindent \textbf{Graphs}. We consider a \textbf{graph} $\matG = (\matV,\matE,\matS)$, where $\matV$ is the set of vertices,
$\matE$ is the set of edges, $\matS$ is the set of raw node features, (e.g. raw text, images), and $\matA \in \mathbb{R}^{|\matV| \times |\matV|}$ is the adjacency matrix.

\noindent \textbf{Node Features $\matX$ from PMs}. 
Denote $\matT$ as a pretained model of any modality, and  $\matX = \matT(\matS)$. $\matX  \in \mathbb{R}^{|\matV| \times d} $ is the extracted $d$-dimensional node feature embedding from the pretrained model $\matT$. We use $\matX \in \mathbb{R}^{|\matV| \times d}$ as features in GNNs. 

\noindent \textbf{Link Prediction}.
Predict whether
there will be a future link $e_{ij}$ between a pair of nodes $i$ and $j$, where $i, j \in \matV$ and $e_{ij} \notin \matE$.

\noindent \textbf{Node Classification}.
Given the labels of $l$ nodes in $\matG$, where $l \ll |\matV|$, predict the unknown classes of $ |\matV|-l$ test nodes. 
\begin{figure}[t!]
	\centering
	\setcounter{subfigure}{0}
	
    \subfloat[h = 1]{%
      \includegraphics[width=0.35\columnwidth]{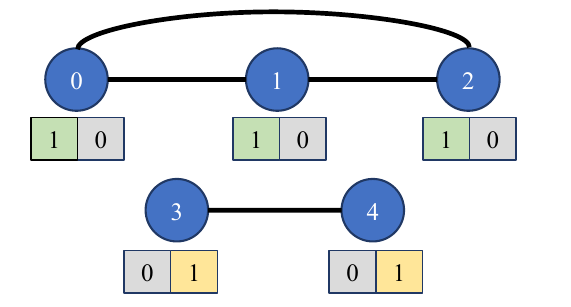}
    }
    ~
    \subfloat[h = -1]{%
      \includegraphics[width=0.35\columnwidth]{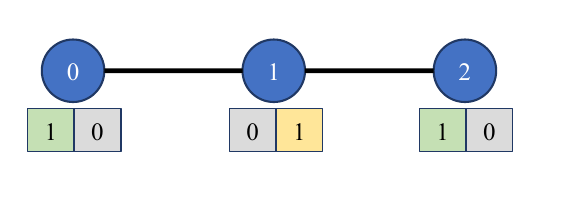}
    }
    ~
\vspace{-0.5cm}
\caption{Example graphs that exhibit (a) strong positive and (b) strong negative correlation between features and structure. (a): 
All linked nodes have the same features;
(b): All linked nodes have complementary features.
}
\vspace{-0.5cm}
    \label{fig:assortativity-example}
\end{figure}
\vspace{-0.3cm}
\subsection{Detection: Proposed Homophily Measure}
\label{sec:homophily}
How can we detect the discrepancy between features and structure? An intuitive approach is to use cosine similarity between nodes or measure the feature smoothness over nodes. However, these methods have several drawbacks, which we outline next.

\noindent \textbf{Intuitive Measures \& their Limitations.}  
The most intuitive way to measure node similarity over connected edges is to measure the \textit{cosine similarity} between nodes, which is not ideal. As shown in Figure \ref{fig:assortativity-example}(b), the graph's cosine similarity is 0 and fails to differentiate random features from negatively correlated features. 

Another candidate measure is \textit{feature smoothness} ~\cite{hou2022measuring}. It is defined as 
$
    \lambda_f = \frac{\left|\left|\sum_{v \in V}(\sum_{v' \in N_v} (\feat_v - \feat_{v'}))^2\right|\right|_1}{|E| \cdot d}
$, where $N_v$ are neighbors of node $v$ and $\lambda_f \in [0, +\infty)$. However, this metric suffers from two drawbacks. First, the magnitude of $\lambda_f$ is unbounded and depends on the number of edges. Thus, feature smoothness cannot be compared across graphs of varying sizes. Second, similar to cosine similarity, feature smoothness fails to differentiate between random and negatively correlated features, as both generate a large $\lambda_f$.

\noindent \textbf{Proposed Feature Homophily Measure.} Motivated by the limitations of the above-mentioned intuitive metrics, we now introduce a new feature homophily measure $h_f$, which can effectively quantify the alignment between the node features and graph structure. 
\vspace{-0.2cm}
\begin{definition}[Feature homophily]
\label{dfn:featureHomophily} Given a graph $\matG = (\matV,\matE)$ and the feature embeddings \textbf{$\feat_i$}, \textbf{$\feat_j$} of nodes i,j where $e_{ij} \in \matE$, the feature homophily ratio $h_f$ is defined as follows:
\vspace{-0.2cm}
\begin{equation}
    h_f = \frac{\sum_{e_{ij} \in E}(\feat_i - \Bar{x}) \cdot (\feat_j - \Bar{x})}{\sqrt{\sum_{e_{ij} \in E}(\feat_i - \Bar{x}) \cdot (\feat_i - \Bar{x})} \cdot \sqrt{\sum_{e_{ij} \in E}(\feat_j - \Bar{x}) \cdot (\feat_j - \Bar{x})}}
    \label{eq:homophily}
\end{equation}
\vspace{-0.3cm}
where $\Bar{x} = \frac{ \sum_{e_{ij}\in E}(\feat_i + \feat_j)}{2|E|} $. 
 \end{definition}

The feature homophily score $h_f$ effectively measures the correlation between nodes over  edges. Its range is confined between $[-1,1]$. In various real-world cases, most graphs exhibit a feature homophily where $h>0$. Figure~\ref{fig:assortativity-example} illustrates that when $h=1$, every connected node pair possesses the \emph{exact same} feature, and when $h=-1$, each connected node pair exhibits the \emph{exact opposite} feature. When $h=0$, the linked nodes showcase random features with no correlation with the graph connectivity.

\noindent \textbf{Difference between Feature Homophily vs. Label Homophily.} In the GNN literature, the most popular notion of homophily is \emph{label} homophily~\cite{zhu2020beyond,ma2021homophily}. While label homophily is related to feature homophily, they are inherently different. Label homophily captures the label similarity between nodes and their immediate neighbors, so it is contingent on the availability of node labels. 
In this work, we tackle different graph learning tasks, including but not limited to link prediction, wherein nodes lack class labels.

\begin{figure}[t!]
    \centering
    \vspace{-0.35cm}
    \includegraphics[width=0.9\linewidth]{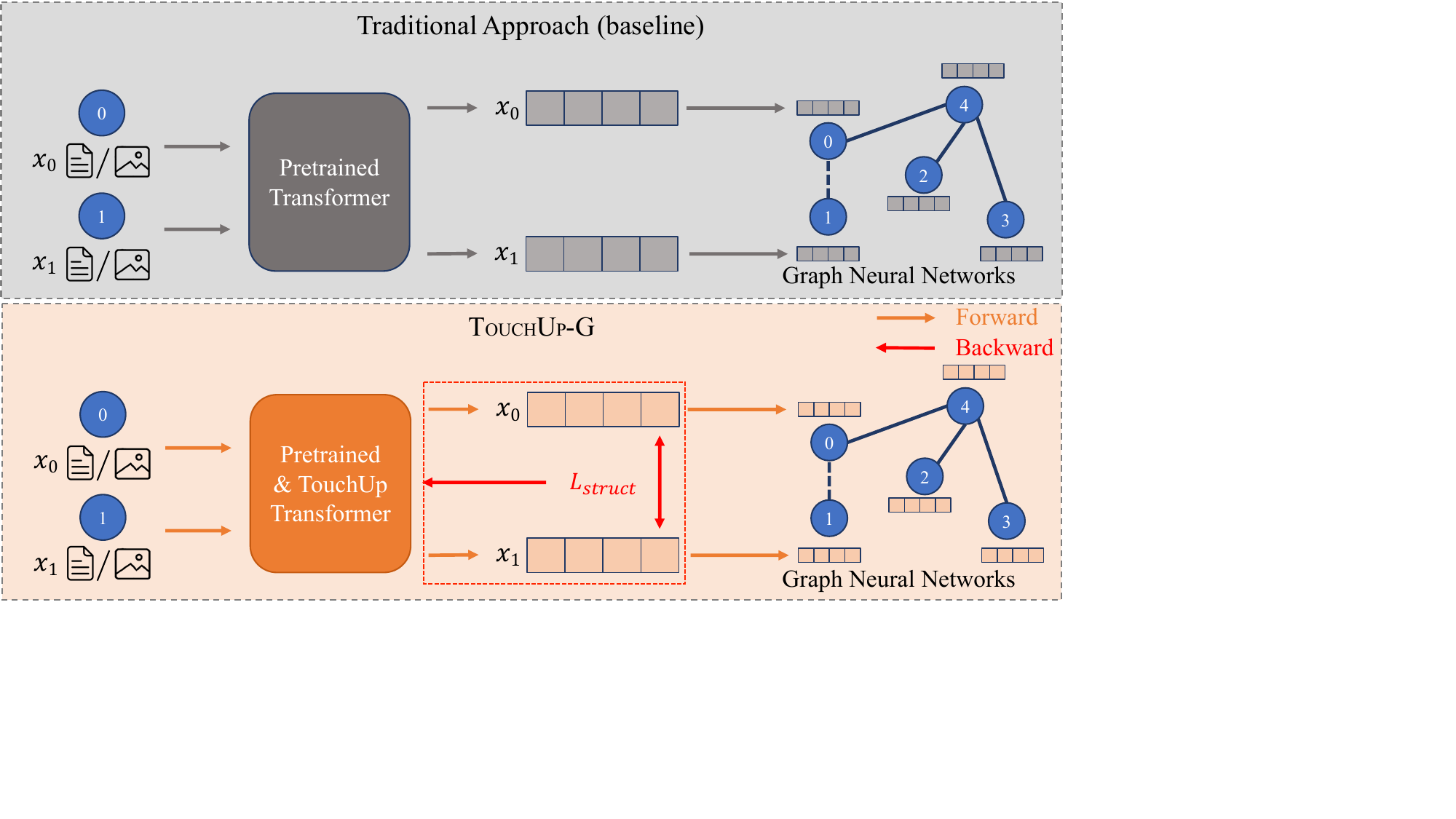}
    \vspace{-0.2cm}
    \caption{\textbf{Overview of \method.} [Top, \tikzsquare[black, fill=msgray]{2.5pt}] A PM~\cite{devlin2018bert,dosovitskiy2020image} is used to extract features from raw text or images, and then GNNs are trained upon the extracted features. [Bottom, \tikzsquare[black, fill=msorange]{2.5pt}] We propose graph-centric finetuning on PMs to correct the discrepancy between features and the  structure.}
    \label{fig:method}
    \vspace{-0.7em}
\end{figure}

\vspace{-0.2cm}
\subsection{Correction: Graph-Centric Finetuning}
We propose graph-centric finetuning, \method,  a simple \textsc{TouchUp} enhancement that refines node features from any PM. \textbf{Given}:

\begin{compactitem}
\item an undirected graph $\matG = (\matV,\matE)$ and its adjacency matrix $\matA \in \mathbb{R}^{|\matV| \times |\matV|}$;
\item   the set $\matS$ of raw node features for all nodes $n \in \matV$; and
\item   a pretrained model $\matT$ that transforms raw node features to node feature embeddings, $\matX = \matT(\matS)$, $\matX \in \mathbb{R}^{|\matV| \times d}$; 
\end{compactitem}
We \textbf{fine-tune} $\matT$ to minimize $L_{\text{struct}}$ using negative sampling:
\vspace{-0.3cm}
\begin{equation}{
   L_{\text{struct}} =  - \frac{1}{|E|} \sum_{\substack{(u,v) \in E}}   ( \log ( \feat_u \cdot \feat_v) + \log  (1-\feat_u \cdot \feat_{v'})
   ),
}
\end{equation}
\vspace{-0.4cm}

where $v' \in \matV$ is a randomly sampled negative and $(u,v) \notin E$. 

An illustrative example is provided in Fig.~\ref{fig:method}. 
We note that, for link prediction, edges used for validation and testing are unobserved, and thus are not used for finetuning.

\noindent \textbf{Non-specific Loss for PMs.} Unlike previous works that hinge on the domain-specific knowledge of text~\cite{jin2023patton}, \method does not rely on prior assumptions about the pretrained models or the source of node features. It can be readily adapted to any PM from any modality.
In our experiments, we  demonstrate that its effectiveness on both text and image-rich graphs.

\section{Experiments}
\label{sec:experiments}

\begin{table}[t!]

\caption{\textbf{Datasets used in \method} \underline{LP} = Link Prediction. \underline{NC} = Node Classification.}
\label{tab:datasets}
\centering
\vspace{-1em}
\resizebox{0.48\textwidth}{!}
{
\begin{tabular}{l@{\hskip5pt}r@{\hskip5pt}r@{\hskip5pt}r@{\hskip5pt}r@{\hskip5pt}r}
\toprule
   \textbf{Name} & \textbf{Nodes} & \textbf{Edges}  & \textbf{Node Features} & \textbf{Pretrained Model} & \textbf{Downstream Task} \\
\midrule 

     \ogbproducts \cite{hu2020open} &  2,449,029 &  61,859,140  &  Text & BERT~\cite{devlin2018bert} & LP \& NC\\ 

    \books~\cite{wan2018item, wan2019fine}  & 1,098,672 & 33,619,434&  Text & BERT~\cite{devlin2018bert} & LP\\

    \amazoncp ~\cite{ni2019justifying} & 379,770 & 4,102,444 & Image & ViT~\cite{dosovitskiy2020image} & LP\\ 
    \ogbarxiv~\cite{hu2020open} & 169,343 & 1,166,243 &  Text & SciBERT~\cite{beltagy2019scibert} & NC\\
\bottomrule
\vspace{-1.2cm}
\end{tabular}

}
\end{table}

We aim to answer the following research questions:

\begin{compactitem}
    \item (\textbf{RQ1}) \multimodal: 
    Can \method handle any modality such as text, images on link prediction?
    \item (\textbf{RQ2}) \general: Can \method
    handle other downstream tasks such as node classification? 
    \item (\textbf{RQ3}) \principled: According to the feature homophily score $h_f$, how correlated are the features from PMs vs.\ the feature from \method?
\end{compactitem}

\enlargethispage{\baselineskip}

\noindent \textbf{Data.} 
We use four public datasets: \ogbproducts, \books, \amazoncp, \ogbarxiv. The details of these datasets are shown in Table~\ref{tab:datasets}. 
\books~ is constructed following~\cite{safavi2022augmenting}. \amazoncp~ is constructed from Amazon-Review~\cite{ni2019justifying} by extracting the co-purchasing links and each product's image as the raw node features. Nodes with missing images or  insufficient density  (i.e., degree$<5$) are eliminated.

\noindent \textbf{Pretrained Models}.
\label{sec:pretrained-model} For \ogbproducts ~and \books~, we use BERT
~\cite{devlin2018bert}.  For \ogbarxiv, we use SciBERT~\cite{beltagy2019scibert}. For BERT and SciBERT, the last layer is dropped when generating representations.
For \amazoncp, we adapt the ImageNet pretrained ViT~\cite{dosovitskiy2020image}. We drop the last layer and add a linear layer to project the embeddings to 256 dimensions.

\noindent \textbf{\method Variants}. 
As shown in Table~\ref{tab:assortativity-score}, we detect that, for all four datasets, $h_f$ is close to 0 when using PMs directly, thus we do correction on all the datasets. We consider two GNN backbones:  SAGE~\cite{hamilton2017inductive} and GATv2~\cite{brody2021attentive}.

\begin{table}[t!]
\centering
\caption{\textbf{\method is multi-modal:}  \method has the best overall performance across all datasets and modalities. We do not report Patton~\cite{jin2023patton} on \amazoncp ~as it only works for text features. OOM = Out of Memory.
}
\vspace{-0.4cm}
\label{tab:text-lp}
\resizebox{.48\textwidth}{!}
{
\begin{tabular}{@{}llcc cc cc@{}}
\toprule
& & \multicolumn{2}{c}{\textbf{SAGE}} && \multicolumn{2}{c}{\textbf{GATv2}}
\\
\cline{3-4} \cline{6-7} 
\textbf{Datasets} &  \textbf{Methods} & \textbf{MRR $\uparrow$} & \textbf{H@10 $\uparrow$} & & \textbf{MRR $\uparrow$} & \textbf{H@10 $\uparrow$} & \\
 \midrule
\multirow{5}{*}{\textbf{\books}} &  \textbf{\degree} &  15.58 $\pm$ 0.60 & 29.45 $\pm$ 0.93 && OOM &  OOM \\
& \textbf{\deepwalk} &  19.52 $\pm$ 0.11 &  30.28 $\pm$  0.20 && OOM &  OOM \\
& \textbf{\bert} &  47.05 $\pm$  0.43 &  73.30 $\pm$  0.38  && OOM &  OOM\\
& \textbf{Patton} &  \underline{65.55 $\pm$  0.08} &  \underline{86.00 $\pm$  0.02} && OOM & OOM \\
& \textbf{\method} & \colorbox{Gray}{ 69.61 $\pm$ 0.38} & \colorbox{Gray}{89.09  $\pm$ 0.61}  && OOM &  OOM  \\
 \midrule
\multirow{5}{*}{\textbf{\ogbproducts}} &  \textbf{\degree} &  10.37 $\pm$ 0.42 & 22.62 $\pm$ 1.16 
&& OOM &  OOM \\

& \textbf{\deepwalk} &  21.14 $\pm$ 0.15 &  32.90 $\pm$  0.06 && OOM & OOM\\
& \textbf{\bert} &  68.65 $\pm$  0.14 &  76.73 $\pm$  0.02 && OOM & OOM\\
& \textbf{Patton} &  \underline{71.70 $\pm$ 0.01} &  \underline{77.66 $\pm$ 0.04} &&  OOM & OOM \\
& \textbf{\method} & \colorbox{Gray}{ 73.66 $\pm$ 0.31} & \colorbox{Gray}{80.40  $\pm$ 0.76} && OOM & OOM \\
\midrule
\multirow4{*}{\textbf{\amazoncp}} &  \textbf{\degree} &  30.03 $\pm$ 0.53 & 74.15 $\pm$ 0.01 && 30.71 $\pm$ 0.82 & 72.89 $\pm$ 0.53 \\
& \textbf{\deepwalk} &  36.05 $\pm$ 0.01 &  96.41 $\pm$ 0.01 && 35.99 $\pm$ 0.31 &  89.57 $\pm$ 0.01 \\
& \textbf{\vit} &  36.08 $\pm$  0.01 &  96.41 $\pm$  0.01 && 40.14 $\pm$ 0.09 &  \colorbox{Gray}{97.81 $\pm$ 0.01}\\
& \textbf{\method} & \colorbox{Gray}{ 41.19 $\pm$ 0.07} & \colorbox{Gray}{97.35  $\pm$ 0.01} && \colorbox{Gray}{42.84 $\pm$ 0.08} &  97.05 $\pm$ 0.07  \\
\bottomrule
\end{tabular}
}
\vspace{-0.6cm}
\end{table}

\noindent \textbf{Metrics.} We evaluate the performance on link prediction and node classification. For link prediction, we report MRR, Hits@10, the two most commonly-used evaluation metrics~\cite{hu2020open,zhu2024pitfalls}. For node classification, we report accuracy, following~\cite{hu2021ogb}. For all evaluation metrics, the higher the number is, the better. We perform hyperparameter tuning using grid search and choose the best performing ones on validation sets. Results are reported on test sets.

\noindent \textbf{Baselines}.  As in~\cite{cui2022positional}, we use \degree ~and \deepwalk ~as embedding-based baselines. For PMs, we mainly consider \bert\cite{devlin2018bert}, and \vit\cite{dosovitskiy2020image}. We also report Patton ~\cite{jin2023patton}, the most recent framework that captures the  dependency between textual attributes and  structure.  For node classification,we compare against \ogb: the original features~\cite{hu2020open}, \bert, \scibert and \deberta~\cite{he2020deberta}. We also report results for two methods that finetune LLMs, GIANT and GLEM~\cite{zhao2022learning,chien2021node}.

\vspace{-0.1cm}
\subsection{(RQ1)  \multimodal: Link Prediction Results}
\label{sec:exp-effective}
\noindent \textbf{Setup \& Evaluation.}
We first evaluate \method's effectiveness on link prediction for feature-rich graphs. Here, we consider text and image---due to the lack of data, we leave other modalities for future study. We report the performance of two widely-used GNN backbones: SAGE and GATv2.

\noindent \textbf{Results.} The quantitative results are shown in Table~\ref{tab:text-lp}. \method achieves state-of-the-art performance at most times with different GNN backbones, even compared with Patton. Especially, there is more than 20$\%$ MRR boost compared with \bert. This indicates that structure-fused  features are more effective than any feature or structure alone. In Fig.~\ref{fig:teaser-qual}, we also provide a qualitative co-purchasing example in \amazoncp ~during test. Given the query "shower head",  our goal is to predict  "bath towel hanger" (ground truth) as a co-purchased item. While \vit \xspace fails, \method correctly predicts the ground truth in the top-2 predictions. Moreover, the "tissue hanger" that \method predicts is also related to bathroom equipment and is likely to be purchased together with shower heads. This qualitative example shows that \method yields more meaningful co-purchasing predictions compared with \vit ~alone, and showcases \method's multimodal ability.
\vspace{-0.3cm}

\begin{figure}[t!]
\centering
\includegraphics[width=0.75\columnwidth]{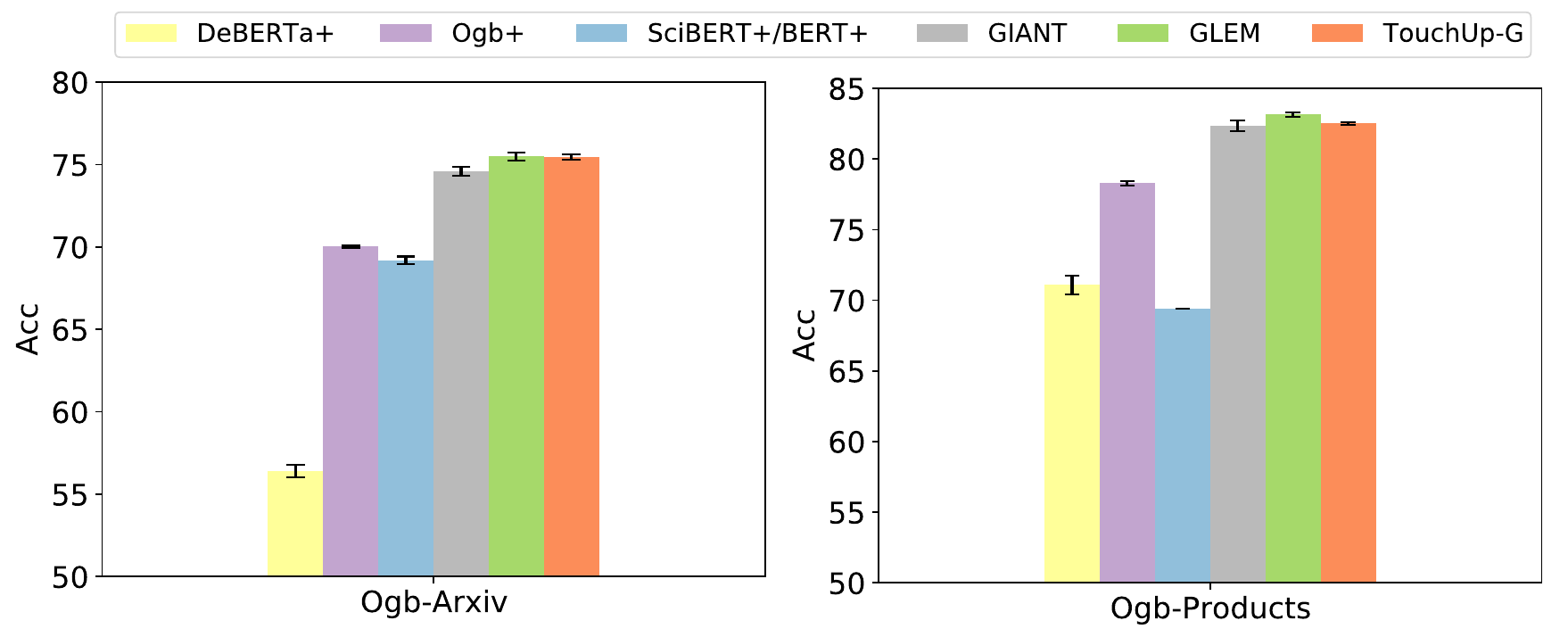}
\vspace{-0.4cm}
\caption{\myTag{\method is \general:} Node classification Results. 
\textit{\method does not use any node label information during training}. However, we obtain comparable performance compared with baselines explicitly finetuned on node labels ~\cite{zhao2022learning}.
}
\vspace{-0.1cm}
    \label{fig:Arxiv}
\end{figure}

\subsection{(RQ2) \general: Node Classification Results}
\begin{table}[t]
\vspace{-0.2cm}
\caption{\myTag{\principled: \method wins.} Feature homophily $h_f$ on various methods. \method gives more than 2$\times$ increase in feature homophily and indicates better correlation
between node features and graph structure. Gain$\uparrow$ denotes the improvements of \method over the same PMs. }
\label{tab:assortativity-score}
\vspace{-0.3cm}
\centering
\resizebox{.44\textwidth}{!}
{
\begin{tabular}{lrrrrr || r r}

\toprule

   Dataset & BERT+~\cite{devlin2018bert} & SciBERT+~\cite{beltagy2019scibert} & ViT+~\cite{dosovitskiy2020image}  & Patton~\cite{jin2023patton} & \textbf{\method} & Gain $\uparrow$\\
\midrule
     \books & 0.137 & - &   - & \textcolor{red}{0.2373} &  \textbf{0.579} \color{green}{(4.2x)} & 47$\%$\\
     \ogbproducts & 0.223 & - &    - &  \textcolor{red}{0.4595} & \textbf{0.762} \color{green}{(3.4x)}  & 7$\%$ \\
     \amazoncp & - & -  & 0.173 & - & \textbf{0.622} \color{green}{(3.6x)} & 14$\%$  \\
     \ogbarxiv & - & 0.194 &  - & - &  \textbf{0.408} \color{green}{(2.1x)} & 9$\%$\\
\bottomrule
\end{tabular}
}
\vspace{-0.6cm}
\end{table}

\noindent \textbf{Setup \& Evaluation.} We report the node classification performance on \ogbarxiv ~and \ogbproducts. Results for \deberta, GIANT, GLEM are directly adapted from~\cite{zhao2022learning}, and the \ogb results are directly adapted from~\cite{wang2019deep}. 

\noindent \textbf{Results.} The results are shown in Fig.~\ref{fig:Arxiv}. \method is comparable  with  GIANT and GLEM. This observation is noteworthy as we do not use any node label information in \method. Our method surpasses all \scibert~ and \bert, indicating the generality of \method across backbones. Moreover, \ogb~ consistently outperforms PM baselines. This further supports our argument that directly utilizing contextualized features from PMs without domain adaptation can negatively impact the performance when the contextualization is irrelevant, and highlights the  necessity of \method.

\vspace{-0.3cm}
\subsection{(RQ3) \principled: \matrices Score}

\noindent \textbf{Setup \& Evaluation.}
For each dataset in Table ~\ref{tab:datasets}, we compute $h_f$ using the representations derived from both the PMs and  \method, before training a GNN.

\noindent \textbf{Results.}
The results are shown in Table~\ref{tab:assortativity-score}.  All datasets exhibit low feature homophily scores prior to fine-tuning. This suggests a lack of alignment between the node features and graph structure across all datasets. However, upon applying \method,  all datasets witness more than 2$\times$ increase in $h_f$, and the state-of-the-art performance on downstream graph tasks. Patton also shows increase in $h_f$ when compared with \bert, which suggests that GNNs' performance can be improved by increasing $h_f$.

\section{Conclusion}
\label{sec:conclusion}
We have presented \method, a simple "Detect \& Correct" approach for refining node features extracted from PMs. \method is \general,  \multimodal, \principled ~and \effective. As future work, we  envision that the finetuning part of \method can be done more efficiently using delta finetuning ~\cite{ding2022delta}.

\balance
\bibliographystyle{ACM-Reference-Format}
\bibliography{BIB/bibliography}

\end{document}